\documentclass[11pt,a4paper]{article}
\PassOptionsToPackage{breaklinks}{hyperref}
\usepackage[hyperref]{emnlp2020}
\usepackage{times}
\usepackage{latexsym}
\usepackage{xspace}

\usepackage{subcaption}
\usepackage{multirow}
\usepackage{array}
\newcolumntype{L}[1]{>{\raggedright\let\newline\\\arraybackslash\hspace{0pt}}p{#1}}
\newcolumntype{C}[1]{>{\centering\let\newline\\\arraybackslash\hspace{0pt}}p{#1}}
\newcolumntype{R}[1]{>{\raggedleft\let\newline\\\arraybackslash\hspace{0pt}}p{#1}}
\usepackage{pbox}
\usepackage{adjustbox}
 
\usepackage[disable]{todonotes}
\usepackage{comment}
\newcommand{\rachel}[1]{\todo[size=\small,color=orange]{\sffamily RB: #1}}

\usepackage[font=small]{caption}

\usepackage{tikz}
\usepackage{pgfplots}
\usepackage{graphicx}
\usepackage{booktabs}
\pgfplotsset{compat=1.14}
\usepackage[utf8]{inputenc}

\usepackage{amsmath}
\usepackage{amssymb}
\usepackage{bm}

\usepackage{lingmacros}
\usepackage{lipsum}
\usepackage{soul}
\usepackage{url}

\aclfinalcopy 

\usepackage{xspace}

\newcommand\bleu{\textsc{Bleu}\xspace}
\newcommand\sentbleu{\textsc{sentBleu}\xspace}
\newcommand\beam{\textsc{Beam}\xspace}
\newcommand\random{\textsc{Random}\xspace}
\newcommand\sampled{\textsc{Sampled}\xspace}
\newcommand\laser{\textsc{Laser}\xspace}

\newcommand\hyter{\texttt{HyTER}\xspace}
\newcommand\treelstm{\textsc{TreeLstm}\xspace}

\newcommand\treelstmplain{\textsc{TreeLstm}\xspace}
\newcommand\human{\textsc{Human}\xspace}
\newcommand\ngram{$n$-gram\xspace}
\newcommand\ngrams{$n$-grams\xspace}

\title{A Study in Improving \bleu Reference Coverage \\with Diverse Automatic Paraphrasing}

\newcommand{\edinburgh}{${}^1$}
\newcommand{\tartu}{${}^2$}
\newcommand{\jhu}{${}^3$}

\author{Rachel Bawden\edinburgh \quad Biao Zhang\edinburgh \quad  Lisa Yankovskaya\tartu \quad Andre Tättar\tartu \quad Matt Post\jhu \\
  {\edinburgh}School of Informatics, University of Edinburgh, Scotland\\
  {\tartu}University of Tartu, Tartu, Estonia \\
  {\jhu}Johns Hopkins University, Baltimore, Maryland, USA
  }

\date{}

\begin{document}
\maketitle
\begin{abstract}
We investigate a long-perceived shortcoming in the typical use of \bleu: its reliance on a single reference.
Using modern neural paraphrasing techniques, 
we study whether automatically generating additional \textit{diverse} references can provide better coverage of the space of valid translations and thereby improve its correlation with human judgments.
Our experiments on the into-English language directions of the WMT19 metrics task (at both the system and sentence level) show that using paraphrased references does generally improve \bleu, and when it does, the more diverse the better. However, we also show that better results could be achieved if those paraphrases were to specifically target the parts of the space most relevant to the MT outputs being evaluated. Moreover, the gains remain slight even when using human paraphrases elicited to maximize diversity, suggesting inherent limitations to \bleu's capacity to correctly exploit multiple references. Surprisingly, we also find that adequacy appears to be less important, as shown by the high results of a strong sampling approach, which even beats human paraphrases when used with sentence-level \bleu.\footnote{Our code and outputs are available at \url{https://github.com/rbawden/paraphrasing-bleu}.}
\end{abstract}

\section{Introduction}\label{sec:intro}

There is rarely a single correct way to translate a sentence; work attempting to encode the entire translation space of a sentence suggests there may be billions of valid translations \citep{dreyer-marcu-2012-hyter}.
Despite this, in machine translation (MT), system outputs are usually evaluated against a single reference.
This especially affects MT's dominant metric, \bleu \citep{papineni-etal-2002-bleu}, since it is a surface metric that operates on explicit $n$-gram overlap 
(see.~(\ref{ex:multiple-translations}) showing two adequate MT outputs, one with only minimal overlap with the reference):\footnote{See~Sec.~\ref{sec:baselines} and \citep[\S1.1]{papineni-etal-2002-bleu} for details.}
\enumsentence{\small
\textit{Ref: This did not bother anybody .}\\
MT$_1$: \textbf{This} didn 't \textbf{bother anybody} .\\
MT$_2$: Nobody was bothered by \textbf{this} .
\label{ex:multiple-translations}}
\vspace{-0.1cm}

Almost since its creation, \bleu's status as the dominant metric for MT evaluation has been challenged (e.g.,~\citet{callison-burch-etal-2006-evaluating}, \citet{mathur-etal-2020-tangled}).
Such work typically uses only a \emph{single} reference, however, which is a deficient form of the metric, since one of \bleu's \emph{raisons d'{\^{e}tre}} was to permit the use of multiple references, 
in a bid to represent ``legitimate differences in word choice and word order.''
Unfortunately, multiple references are rarely available due to the high cost and effort of producing them.
One way to inexpensively create them  is with automatic paraphrasing.
This has been tried before \citep{zhou-etal-2006-evaluating,kauchak-barzilay-2006-paraphrasing}, but only recently have paraphrase systems become good enough to generate fluent, high quality \emph{sentential} paraphrases (with neural MT-style systems). 
Moreover, it is currently unclear (i)~whether adding automatically paraphrased references can provide the diversity needed to better cover the translation space, and (ii) whether this increased coverage overlaps with observed and valid MT outputs, in turn improving \bleu's correlation with human judgments.

We explore these questions, 
testing on all into-English directions of the WMT19 metrics shared task \citep{ma-etal-2019-results} at the system and segment level.
We compare two approaches: (i)~generating diverse references with the hope of covering as much of the valid translation space as possible, and (ii)~more directly targeting the relevant areas of the translation space by generating paraphrases that contain \ngrams selected from the system outputs.
This allows us to compare the effects of diversity against an upper bound that has good coverage.
We anchor our study by comparing automatically produced references against human-produced ones on a subset of our data.

Our experiments show that adding paraphrased references rarely hurts \bleu and can provide moderate gains in its correlation with human judgments.
Where it does help, the gains are correlated with diversity (and less so adequacy), but see diminishing returns, and fall short of the non-diverse method designed just to increase coverage. Manual paraphrasing does give the best system-level \bleu results, but even these gains are relatively limited, suggesting that diversity alone has its limits in addressing weaknesses of surface-based evaluation metrics like \bleu.
%
%

\section{Related Work}\label{sec:related-work}

\paragraph{Paraphrasing for MT evaluation}

There is a long history of using paraphrasing to overcome the limitations of \bleu-style metrics. 
Some early approaches rely on external resources (e.g.~WordNet) to
provide support for synonym matching
\citep{banerjee-lavie-2005-meteor,kauchak-barzilay-2006-paraphrasing,denkowski:lavie:meteor-wmt:2014}. More
automatic methods of identifying paraphrases have also been developed.
An early example is ParaEval \citep{zhou-etal-2006-evaluating}, which
provides local paraphrase support using paraphrase sets automatically
extracted from MT phrase tables.  More recently,
\citet{apidianaki-etal-2018-automated} exploit contextual word
embeddings to build automatic \hyter networks. However they achieve
mixed results, particularly when evaluating high performing (neural)
models.

The use of MT systems to produce paraphrases has also been studied previously. 
\citet{albrecht-hwa-2008-role} create pseudo-references by using
out-of-the-box MT systems 
and see improved correlations with human judgments, helped by
the systems being of better quality than those evaluated. 
This method was extended by
\citet{yoshimura-etal-2019-filtering}, who filter the
pseudo-references for quality.  An alternative strategy is to use
MT-style systems as paraphrasers, applied to the
references. \citet{madnani-etal-2007-using} show that additional
(paraphrased) references, even noisy ones, reduce the number of human
references needed to tune an SMT system, without significantly
affecting MT quality. However their aim for coverage over quality
means that their paraphrases are unlikely to be good enough for use in
a final evaluation metric.

Despite the attention afforded to the task, success has been limited
by the fact that until recently, there were no good
\textit{sentence-level} paraphrasers---\citet{federmann-etal-2019-multilingual} showed that neural
paraphrasers can now outperform humans for adequacy and cost.
Attempts (e.g.~\citealp{napoles-etal-2016-sentential}) using earlier
MT paradigms were not able to produce fluent output, and publicly
available paraphrase datasets have only been recently released
\citep{wieting-gimpel-2018-paranmt,hu-etal-2019-improved}. Moreover,
most works focus on synonym substitution rather than more radical
changes in sentence structure, limiting the coverage
achieved.

\paragraph{Structurally diverse outputs}
Diverse generation is important to ensure a wide coverage of possible
translations. Diversity, both lexical and structural, has been a major
concern of text generation tasks
\citep{colin-gardent-2018-generating,iyyer-etal-2018-adversarial}.
State-of-the-art neural MT-style text generation models used for
paraphrasing
\citep{prakash-etal-2016-neural,mallinson-etal-2017-paraphrasing}
typically suffer from limited diversity in the beam. Techniques such
as sampling from the model distribution or from noisy outputs have
been proposed to tackle this \citep{edunov-etal-2018-understanding}
but can harm output quality.

An effective strategy to encourage structural diversity is to add
syntactic information (which can be varied) to the generated text. The
constraints can be specified manually, for example by adding a parse
tree \citep{colin-gardent-2018-generating,
iyyer-etal-2018-adversarial} or by specifying more abstract
constraints such as rewriting embeddings \citep{xu_Dpage_2018}. A
similar but more flexible approach was adopted more recently by
\citet{shu-etal-2019-generating}, who augment target training
sentences with cluster pseudo-tokens representing the structural
signature of the output sentence. When decoding, the top cluster codes
are selected automatically using beam search and for each one a
different hypothesis is selected. We adopt
\citeauthor{shu-etal-2019-generating}'s approach here, due to the
automatic nature of constraint selection and the flexibility afforded
by constraint definition, allowing us to test different types of
diversity by varying the type of sentence clustering method.

\section{Generating paraphrased references}\label{sec:paraphrasing-methods}

We look at two ways to produce paraphrases of English references using
English--English NMT architectures.  The first
(Sec.~\ref{sec:diverse-paraphrasing}) aims for maximal \textit{lexical
  and syntactic diversity}, in a bid to better cover the space of
valid translations.
In contrast, the second (Sec.~\ref{sec:contrained-paraphrasing}) aims
to produce paraphrases that target the most relevant areas of the
space (i.e.~that are as close to the good system outputs as possible).
Of course, not all outputs are good, so we attempt to achieve coverage
while maintaining adequacy to the original reference by using
information from the MT outputs.  While less realistic practically,
this approach furthers the study of the relationship between diversity
and valid coverage.

\subsection{Creating diverse paraphrases}\label{sec:diverse-paraphrasing}
 
To encourage diverse paraphrases, we use
\citeauthor{shu-etal-2019-generating}'s
(\citeyear{shu-etal-2019-generating}) method for diverse MT, which
consists in clustering sentences according to their type and training
a model to produce outputs corresponding to each type.
Applied to our paraphrasing scenario, the methodology is as follows:
\begin{enumerate}
    \item Cluster target sentences by some property (e.g.,~semantic, syntactic representation);
    \item Assign a code to each cluster and prefix each target sentence in the training data with its code (a pseudo-token), as follows: 
    \enumsentence{\label{ex:prepend}
    \small
    $\langle$cl\_14$\rangle$ They knew it was dangerous .\\
    $\langle$cl\_101$\rangle$ They had chickens, too .\\
    $\langle$cl\_247$\rangle$ That 's the problem .}\vspace{-0.2cm}
    \item Train an NMT-style paraphrase model using this augmented data;
    \item At test time, apply the paraphraser to each reference in the test set; beam search is run for each of the $n$ most probable sentence codes to produce $n$ paraphrases per reference.
\end{enumerate}

As in \citep{shu-etal-2019-generating}, we test two different types of
diversity: \textit{semantic} using \laser sentential embeddings
\citep{doi:10.1162/tacl_a_00288} and \textit{syntactic} using a
TreeLSTM encoder \citep{tai-etal-2015-improved}. Both methods encode
each sentence as a vector, and the vectors are clustered using
$k$-means into 256 clusters (full details in
App.~\ref{app:cluster-training-details}).

\paragraph{Semantic:} 
We use pretrained \laser sentential embeddings
\cite{doi:10.1162/tacl_a_00288} to encode sentences into
1024-dimensional
vectors. 

\paragraph{Syntactic:} 
As in \citep{shu-etal-2019-generating}, we encode constituency trees
into hidden vectors using a TreeLSTM-based recursive autoencoder, with
the difference that we use $k$-means clustering to make the method
more comparable to the above, and we encode syntactic information
only.

\subsection{Output-guided constrained paraphrases}\label{sec:contrained-paraphrasing}

Diversity is good, but even a highly diverse set of references may not
necessarily be in the same space as the MT outputs.
We attempt to achieve high coverage of the system outputs by using a
weak signal from those outputs.  The signal we use is \emph{unrewarded
  \ngrams} from the best systems, which are \ngrams in system outputs
absent from the original reference.  We identify them as follows.  For
each sentence in a test set, we find all \ngrams that are (a)~not in
the reference but (b)~are present in at least 75\% of the system
outputs, (c)~limited to the top half of systems in the human
system-level evaluation \cite{barrault-etal-2019-findings}.  Then, for
each such \ngram, we generate one paraphrase of the reference using
constrained decoding \citep{post-vilar-2018-fast}, with that \ngram as
a constraint.  This gives a variable-sized set of paraphrased
references for each sentence.  In order to limit overfitting to the
best systems, we use a cross-validation framework, in which we
randomly split the submitted systems into two groups, the first used
to compute the \ngram constraints and the augmented references, and
the second half for evaluation.  We repeat this ten times and report
the average correlation across the splits.

\section{Experiments}\label{sec:experiments}

Our goal is to assess whether we can generate paraphrases that are representative of the translation space and which, when used with \bleu, improve its utility as a metric.
We therefore carry out experiments to (i)~evaluate the adequacy and diversity of our paraphrases (Sec.~\ref{sec:eval_diversity}) and (ii)~compare the usefulness of all
methods in improving \bleu's correlation with human judgments of MT quality (Sec.~\ref{sec:exp-metric_eval}).
\bleu is a corpus-level metric, and our primary evaluation is therefore its system-level correlation. 
However, it is often also used at the segment level (with smoothing to avoid zero counts).
It stands to reason that multiple references would be \emph{more} important at the segment-level, so we also look into the effects of adding paraphrase references for \sentbleu too.

\subsection{Metric evaluation}\label{sec:exp-metric_eval}
For each set of extra references, we produce multi-reference \bleu and \sentbleu metrics, which we use to score all into-English system outputs from the WMT19 news task.\footnote{\url{http://statmt.org/wmt19/results.html}.} We evaluate the scores as in the metrics task \citep{ma-etal-2019-results},  by calculating the correlation with manual direct assessments (DA) of MT quality \citep{graham-etal-2013-continuous}.
System-level scores are evaluated using Pearson's $r$ 
and statistical significance of improvements (against single-reference \bleu) using the Williams test \citep{williams_1959}.  
Segment-level correlations are calculated using Kendall's $\tau$ (and significance against single-reference \sentbleu with bootstrap resampling) on the DA assessments transformed into relative rankings.

\subsection{Baseline and contrastive systems}\label{sec:baselines}

Our true baselines are 
case-sensitive corpus \bleu and \sentbleu, both calculated using sacre\bleu \citep{post-2018-call} using the standard \bleu formula.
Though likely familiar to the reader, we review it here. 
\bleu is computed by averaging modified $n$-gram precisions ($p_n$, $n = 1..4$) and multiplying this product by a brevity penalty (BP), which penalizes overly short translations and thereby works to balance precision with recall:
\begin{align}
\text{\bleu} &= \text{BP} \cdot \exp\left(\sum_{n=1}^N w_n \log p_n \right)
\\
\text{BP} &= 
\begin{cases}
    1 \quad\quad\quad\text{ if } c > r\\
    e^{1-r/c} \quad\text{ if } c \leq r 
\end{cases} \\
p_n &= \dfrac{\sum_{h\in H} \sum_{\text{ngram}\in h} \#_{\text{clip}}\left(\textit{ngram}\right)}{\sum_{h'\in H} \sum_{\text{ngram'}\in h'} \#\left(\textit{ngram'}\right)},
\label{eq:ngramprecisions}
\end{align}
with $c$ and $r$ the lengths of the hypothesis and reference sets respectively, $H$ is the set of hypothesis translations, $\#\left(\textit{ngram}\right)$ the number of times $\textit{ngram}$ appears in the hypothesis, and \#$_{\text{clip}}(\textit{ngram})$ is the same but clipped to the maximum number of times it appears in any one reference.

By definition, \bleu is a corpus-level metric, since the statistics above are computed across sentences over an entire test set.
The sentence-level variant requires a smoothing strategy to counteract the effect of 0 $n$-gram precisions, which are more probable with shorter texts.
We use exponential smoothing.
Both baselines use the single provided reference only.
We also compare against several contrastive paraphrasing approaches: (i)~\textsc{beam}, which adds to the provided reference the the $n$-best hypotheses in the beam of a baseline paraphraser, and (ii)~\textsc{sampled}, which samples from the top 80\% of the probability mass at each time step \citep{edunov-etal-2018-understanding}. For the sentence encoding methods, we also include (iii)~\textsc{random}, where randomly selected cluster codes are used at training and test time.

As a topline, we compare against manually paraphrased references (\human), which we produce for a subset of 500 sentences from the de--en test set.
Two native English speakers together produced 
five paraphrases per reference (alternately two or three paraphrases).
They were instructed to craft paraphrases that were maximally different (lexically and syntactically) from both the reference and the other paraphrases (to which they had access), without altering the original meaning.

\subsection{Paraphrase model training}\label{sec:model-training}
We train our paraphrasers using data from Parabank~2~\cite{hu-etal-2019-large},  containing $\approx$20M sentences with up to 5 paraphrases each, of which we use the first paraphrase only.
We preprocess by removing duplicate sentences and those longer than 100 words and then segment into subwords using SentencePiece \cite{kudo-richardson-2018-sentencepiece} (unigram model \cite{kudo-2018-subword} of size 16k).
The data splits are created by randomly shuffling the data and reserving 3k pairs each for dev and test.
For syntactic sentence encoding methods,
we use the Berkeley Parser \cite{petrov-etal-2006-learning} (internal tokenisation and prioritizing accuracy)
and prune trees to a depth of 4 for $\approx$6M distinct trees.\footnote{Cf.~App.~\ref{app:parse-depths} for the number of trees at different depths.}

Paraphrase models are Transformer base models \citep{vaswani_attention_2017} (Cf.~App.~\ref{app:training-details} for details).
All models are trained using the Marian NMT toolkit \citep{junczys-dowmunt-etal-2018-marian}, except for \sampled and the constraint approach, for which we use the Sockeye toolkit \cite{hieber-etal-2018-sockeye}, since Marian does not support these features.

For baseline models, we produce $n$ additional references by taking the $n$-best in the beam (using a beam size of 20, which is the maximum number of additional references we test).
For models using cluster codes, paraphrases are produced by selecting the $n$-best cluster codes at the first decoding step and then decoding each of these hypotheses using separate beam searches (of size 6).

\section{Paraphrase Adequacy and Diversity}
\label{sec:results-diversity}

\subsection{Adequacy}

\begin{table*}[!ht]
\centering\small
\resizebox{\textwidth}{!}{
\begin{tabular}{lrp{11.4cm}p{4.8cm}}
\toprule
Reference & DA & \textit{What provoked Lindsay Lohan to such very strange actions is currently completely unclear.} & \textit{Now they have come to an agreement.} \\
\midrule
\multirow{3}{*}{\beam}  & \multirow{3}{*}{91.7}
     & What caused Lindsay Lohan to do such strange things is not clear at the moment. & Now they've made a deal.\\
     && What provoked Lindsay Lohan's strange actions is not clear at the moment. & Now they've reached a deal.\\
     && What has provoked Lindsay Lohan's strange actions is not clear at the moment. & Now they made a deal.\\
\midrule
\multirow{3}{*}{\sampled}  & \multirow{3}{*}{85.0}
     & What prompted Lindsay Lohan's most extraordinary actions? & And now they've agreed.\\
     && What made Lindsay Lohan act so weird? & And now they've agreed. \\
     && What inspired Lindsay Lohan to do such odd things? & They've reached an agreement.\\
\midrule
\multirow{3}{*}{\laser} & \multirow{3}{*}{90.1}
     & What provoked Lindsay Lohan to act so strangely is not clear at the moment. & Now they've reached a deal.\\
     && It's not clear what provoked Lindsay Lohan to act so strangely. & Now they've agreed.\\
     && It's not clear what prompted Lindsay Lohan to act so strangely.& Now they've agreed \\
\midrule
\multirow{3}{*}{\treelstmplain} & \multirow{3}{*}{88.0}
     & What provoked Lindsay Lohan to do such a strange thing is not clear at the moment. & Now they made a deal.\\
     && It is not clear at this time what provoked Lindsay Lohan to do such strange things. & Now they've made a deal.\\
     && The reason that Lindsay Lohan has been provoked by these very strange actions is not clear at the moment. & They've already made a deal.\\
\midrule
\multirow{3}{*}{\human} & \multirow{3}{*}{95.2}
     & It is currently totally unclear what made Lindsay Lohan do such strange things. & They have now come to an agreement.\\
     && The cause of Lindsay Lohan's strange actions is really not clear at the moment. & An agreement has now been made. \\
     && The reasons behind Lindsay Lohan's such bizarre acts are completely obscure for now. & They have reached an agreement.\\
\bottomrule
\end{tabular}
}
\caption{Direct assessment (DA) adequacy scores for the \beam and \sampled baseline, the two diverse approaches and human paraphrases for the 100-sentence de--en subset. We also provide each method's top 3 paraphrases for two references.}
\label{tab:examples}
\end{table*}
To ensure our automatically produced paraphrases are of sufficient quality, we first assess their adequacy (i.e.,~faithfulness to the original meaning).
We determine adequacy by manually evaluating  paraphrases of the first 100 sentences of the de--en test set. We compare a subset of the automatic methods (\beam, \sampled, \laser, \treelstmplain) as well as \human.
5 annotators (2 native and 3 fluent English speakers) rated the paraphrases' adequacy using DA, 
indicating how well (0--100) the official reference's meaning is preserved by its paraphrases.
%
25 judgments were collected per sentence (sampling from each system's top 5 paraphrases)
System-level scores are produced by averaging  across all annotations.

The results and examples of some of the paraphrased references are
given in Tab.~\ref{tab:examples} (more examples are given in
App.~\ref{app:more-examples}). Whilst the task is inherently
subjective, we see a clear preference for human paraphrases, providing
a reference point for interpreting the scores.  The automatic
paraphrase systems are not far behind, and the scores are further
corroborated by the lowest score being assigned to the sampled output,
which we expect to be less faithful to the reference meaning.

\subsection{Diversity}\label{sec:eval_diversity}

We evaluate the diversity of paraphrased references using two diversity scores (DS):

\vspace{3pt}\noindent{\small%
\[ \text{DS$_{\text{x}}$} = \dfrac{1}{|Y|(|Y|-1)}\sum_{y\in Y}\sum_{y'\in Y, y'\neq y} 1 - \Delta_\text{x}\left(y, y'\right),\]%
}

\vspace{-3pt}\noindent where $Y$ is the set of paraphrases of a
sentence produced by a given system, and $\Delta_\text{x}$ calculates
the similarity of paraphrases $y$ and $y'$. We use two different
functions: $\Delta_{BOW}$ (for lexical similarity) and $\Delta_{tree}$
(for syntactic similarity). Both give scores between 1 (identical) and
0 (maximally diverse).

\paragraph{DS$_\textit{BOW}$} is the lexical overlap between the sets of words in two paraphrases. $\Delta_\textit{BOW}(y, y')$ corresponds to the number of unique words in common between $y$ and $y'$, divided by their mean length.

\paragraph{DS$_\textit{tree}$} uses $\Delta_{tree}$, the average tree kernel similarity score between paraphrases.
We compute tree kernels using the ``subset tree'' (SST) comparison tree kernel similarity function presented in \citep[\S2.2]{moschitti-2006-making}, with a decay value 
of $\lambda=0.5$, and excluding leaves ($\sigma=0$).

\begin{table}[!ht]

\centering\small
\resizebox{0.87\linewidth}{!}{
\begin{tabular}{llrrr}
\toprule
$n$ & Method & DS$_{\textit{BOW}}$ & DS$_{\textit{tree}}$ & \bleu \\
\midrule
0 & none & - & - & 29.8 \\
\midrule
\multirow{5}{*}{5} 
& \random & 0.10 & 0.01 & 34.8 \\
& \beam & 0.22 & 0.30 & 37.0  \\
& \laser & 0.24 & 0.33 & 37.5 \\
& \treelstmplain & 0.28 & 0.47 & 37.7 \\
& \sampled & \textbf{0.41} & \textbf{0.56} & 40.1  \\
\midrule
\multirow{3}{*}{5*} 
&  \sampled & 0.40 & 0.55 & 47.0\\
& Constraints & 0.19 & 0.30 & \textbf{56.5} \\
& \textbf{\human} & \textbf{0.80} & \textbf{0.68} & 48.9 \\
\midrule
\multirow{5}{*}{20} 
& \random & 0.10 & 0.01 & 34.8 \\
& \beam & 0.27 & 0.37 & 39.7 \\
& \laser & 0.31 & 0.45 & 41.3 \\
& \treelstmplain & 0.32 & 0.53 & 41.0 \\
& \sampled & \textbf{0.51} & \textbf{0.65} & \textbf{47.3} \\
\midrule
$\infty$ & Constraints & 0.21 & 0.28 & 46.4 \\
\midrule
& MT submissions & 0.37 & 0.51 & - \\
\bottomrule
\end{tabular}}
\caption{Diversity scores (DS) of paraphrased references averaged over all into-English test sets, where $n$ is the number of paraphrases.
The final row indicates diversity among MT outputs. * indicates results just for the 500-sentence de--en subset. The final column is the average \bleu score.
}
\label{tab:paraphrase_diversity-total-ave}

\end{table}

The results (Tab.~\ref{tab:paraphrase_diversity-total-ave}) show that all methods other than \random give more diversity than \beam. 
\citeauthor{shu-etal-2019-generating}'s cluster code method generates diverse paraphrases. 
As expected, random cluster codes are not helpful, producing mostly identical paraphrases differing only in the cluster code.
Diversity increases for all methods as paraphrases are added.
\treelstmplain produces structurally more diverse paraphrases than
\laser and has high lexical diversity too, despite codes being
entirely syntactic, suggesting that structural diversity leads to
varied lexical choices.  The most lexically and structurally diverse
method (except for \human), is in fact the strong baseline \sampled,
which is likely due to the noise added with the method.

The increased diversity is generally reflected by an increase in the
average \bleu score (final column of
Tab.~\ref{tab:paraphrase_diversity-total-ave}). These higher \bleu
scores indicate that the additional paraphrases are better covering
the translation space of the MT outputs, but it remains to be seen
whether this concerns the space of valid and/or invalid
translations. In contrast, some of the diversity makes less of an
impact on the \bleu score; the gap in syntactic diversity between
\laser and \treelstm (+20 references) is not reflected in a similar
gap in \bleu score, indicating that this added diversity is not
relevant to the evaluation of these specific MT outputs.

\section{Metric Correlation Results}
\label{sec:results-correlation}

\begin{table*}[t]
\centering\small
\begin{subtable}{0.75\linewidth} 
\centering\small
\scalebox{0.95}{
\begin{tabular}{llrrrrrrrr}
\toprule
&&& \multicolumn{3}{c}{System Gains} && \multicolumn{3}{c}{Segment Gains} \\
Approach & Method & \hphantom{oo} & Ave. & Min & Max & \hphantom{oo} & Ave. & Min & Max \\
\midrule
\multirow{3}{*}{\pbox{1.8cm}{Baselines (+5)}}
& \beam &&  0.020 & {-0.006} & \textbf{0.059}  &&  0.013 & {-0.001} & {0.029} \\
& \random &&  0.017 & {0.000} & \textbf{0.046}  &&  0.007 & {-0.002} & {0.017} \\
& \sampled &&  0.024 & {-0.002} & \textbf{0.067}  &&  0.017 & {-0.004} & \textbf{0.044} \\
\midrule
\multirow{3}{*}{\pbox{1.8cm}{Diversity (+1)}}
& \laser &&  0.017 & {-0.000} & \textbf{0.048}  &&  0.009 & {-0.003} & {0.025} \\
& \treelstm &&  0.017 & {-0.000} & \textbf{0.048}  &&  0.011 & {-0.002} & {0.027} \\
\midrule
\multirow{2}{*}{\pbox{1.8cm}{Diversity (+5)}}
 & \laser &&  0.020 & {-0.004} & \textbf{0.056}  &&  0.011 & {-0.002} & {0.033} \\
 & \treelstm &&  0.020 & {-0.004} & \textbf{0.057}  &&  0.013 & {-0.004} & {0.030} \\
\midrule
\multirow{2}{*}{\pbox{1.8cm}{Output-specific (+1)}}
& \laser &&  0.012 & {-0.006} & \textbf{0.041}  &&  0.006 & {-0.001} & {0.016} \\
& \treelstm &&  0.014 & {-0.007} & \textbf{0.041}  &&  0.007 & {-0.005} & {0.016} \\
\midrule
Constraints & 4-grams  && {0.025} & -0.002 & \textbf{0.061} && 0.002 & -0.097 & \textbf{0.072} \\
\midrule
Human && & - & - & -  & & - & - & -  \\
WMT-19 best & Multiple && {0.079} & {0.010} & \textbf{0.194} && {0.117} & \textbf{0.072} & \textbf{0.145} \\ 
\bottomrule
\end{tabular}}
\caption{\label{tab:correl-all}Average and minimum and maximum gains over all into-English test sets}
\end{subtable}
\begin{subtable}{0.19\linewidth}
\small
\scalebox{0.95}{
\begin{tabular}{rr}
\toprule
System & Segment \\
\multicolumn{2}{c}{de--en} \\
\midrule
\textbf{0.040}   &  {0.021}  \\
\textbf{0.031}   &  {0.017}  \\
\textbf{0.044}   &  {0.043}  \\
\midrule
\textbf{0.034}   &  {0.022}  \\
\textbf{0.031}   &  {0.011}  \\
\midrule
\textbf{0.040}   &  {0.022}  \\
\textbf{0.044}   &  {0.008}  \\
\midrule
\textbf{0.032}   &  {0.015}  \\
\textbf{0.039}   &  {0.011}  \\
\midrule
-0.027 & 0.035 \\
\midrule
\textbf{0.039}   &  {0.037}  \\
 - & - \\
\bottomrule
\end{tabular}}
\caption{\label{tab:correl-500-subset}500-sample subset}
\end{subtable}
\caption{\label{tab:summary-correlations}Absolute gains in correlation (with respect to the true \bleu and sentence\bleu baseline correlations). Significant gains (except for averages) are marked in bold ($p\leq0.05$). Full results per language pair are provided in App.~\ref{app:full-wmt19-results}. WMT-19 best refers to the best metric scores from the official shared task (the best metric can be different for each language pair).}
\end{table*}


The correlation results for each of the metrics (both system- and
segment-level) for different numbers of additional
references\footnote{The table only reports up to 5 paraphrases; adding
  10 or 20 did not improve any of the correlations further.}
(aggregated full results) are shown in Tab.~\ref{tab:correl-all} and
Tab.~\ref{tab:correl-500-subset} (for the de--en 500-sample
subset). We aggregate the main results to make them easier to
interpret by averaging over all into-English test sets (the
\textit{Ave.} column) and we also provide the gains for the language
pairs that gave the smallest and greatest gains (\textit{Min} and
\textit{Max} respectively).  Full raw results can be found in
App.~\ref{app:full-wmt19-results}.

\paragraph{System-level}  Adding paraphrased references does not significantly hurt performance, and usually improves it; we see small gains for most languages (Ave.\ column), although the size of the gain varies, and correlations for two directions (fi--en and gu--en) are degraded but non-significantly (shown by the small negative minimum 
gains).  

\begin{figure}[!t]
\resizebox{\linewidth}{!}{
\pgfplotsset{
  every axis plot/.append style={line width=0.9pt}
}
\begin{tikzpicture}

	\begin{axis}[
		height=6.6cm,
		width=\linewidth,
		grid=major,
		xmin=0,
		xmax=0.5,
		ylabel = Absolute gain in Pearson's $r$,
		xtick distance = 0.1,
		legend style={
            at={(1,-0.2)},
            legend columns=3,
            draw=none},
        axis x line*=bottom,
        axis y line*=left,
        legend style={draw=none,
            at={(1.4,1), font=\small,
            /tikz/every even column/.append style={column sep=0.9cm}},
    },
    legend columns=1
	]

\addlegendentry{de-en}
\addplot[draw=none,mark=triangle,blue] coordinates {
(0.185709145288,0.031)
(0.0746543504557,0.026)
(0.364103919534,0.036)
(0.192145472612,0.035)
(0.239330620017,0.036)
(0.216712020526,0.038)
(0.0741966857501,0.026)
(0.406292698522,0.047)
(0.230737241657,0.044)
(0.269963525226,0.043)
(0.239071851112,0.044)
(0.0742807581059,0.026)
(0.449424123242,0.052)
(0.264853692762,0.048)
(0.292012329416,0.048)
(0.265442506004,0.046)
(0.0743441329959,0.026)
(0.513844607653,0.056)
(0.305316505046,0.052)
(0.312841386936,0.051)
};

\addlegendentry{fi-en}
\addplot[draw=none,mark=square,red] coordinates {
(0.179380070839,-0.001)
(0.067224397588,0.001)
(0.352168930985,0.001)
(0.186568123227,0.001)
(0.239742247385,-0.001)
(0.214855945195,-0.001)
(0.0667466038154,0.001)
(0.398914866529,-0.001)
(0.223827903416,0.0)
(0.270567680591,-0.003)
(0.237574420991,-0.001)
(0.0667741075469,0.001)
(0.442896972763,-0.005)
(0.261719190829,-0.004)
(0.294236144162,-0.005)
(0.263789441628,0.0)
(0.0668371802924,0.001)
(0.510841138874,-0.009)
(0.304435572691,-0.005)
(0.315291790801,-0.005)
};

\addlegendentry{gu-en}
\addplot[draw=none,mark=o,black] coordinates {
(0.181524524533,-0.005)
(0.0645037837645,0.006)
(0.320859016606,0.0)
(0.181932641431,0.0)
(0.249056935588,-0.001)
(0.214016092374,-0.006)
(0.0651720400453,0.006)
(0.372088694688,-0.001)
(0.215881014343,-0.004)
(0.270465321443,-0.003)
(0.238584002409,-0.006)
(0.0654352418899,0.006)
(0.417351869903,0.001)
(0.253755360147,-0.011)
(0.292779656384,-0.006)
(0.267272126295,-0.003)
(0.0654990100256,0.006)
(0.485675344052,0.001)
(0.295277322229,-0.028)
(0.31623493059,-0.01)
};

\addlegendentry{kk-en}
\addplot[draw=none,mark=*,gray] coordinates {
(0.19394690794,0.015)
(0.0901129692546,0.014)
(0.35714718999,0.019)
(0.200729967661,0.016)
(0.266323372536,0.017)
(0.226335427159,0.018)
(0.0905515814413,0.014)
(0.410110409491,0.023)
(0.241781899105,0.02)
(0.291589039057,0.021)
(0.250249134506,0.02)
(0.0902164072461,0.014)
(0.454733560258,0.027)
(0.274867126437,0.024)
(0.31337100985,0.024)
(0.275842935874,0.022)
(0.0899433525163,0.014)
(0.517551855,0.03)
(0.314518868276,0.027)
(0.333436608627,0.026)
};

\addlegendentry{lt-en}
\addplot[draw=none,mark=square*,olive] coordinates {
(0.192409268255,0.016)
(0.114189104179,0.014)
(0.38090873517,0.018)
(0.211635964415,0.015)
(0.250666208805,0.015)
(0.223546688363,0.017)
(0.114874816275,0.014)
(0.418017365479,0.02)
(0.238271561085,0.018)
(0.276328809372,0.018)
(0.243013833544,0.017)
(0.115147556036,0.014)
(0.45267081486,0.023)
(0.26817711884,0.019)
(0.295758933091,0.019)
(0.263703756824,0.018)
(0.115382821325,0.015)
(0.507564808455,0.023)
(0.304980762987,0.02)
(0.313283537107,0.02)
};

\addlegendentry{ru-en}
\addplot[draw=none,mark=diamond,orange] coordinates {
(0.192356388546,0.052)
(0.0893812728579,0.045)
(0.357841406864,0.055)
(0.202511878767,0.05)
(0.251704295321,0.053)
(0.222860223282,0.059)
(0.089680914947,0.046)
(0.401354205522,0.067)
(0.237501374569,0.056)
(0.275047806255,0.056)
(0.243728566391,0.062)
(0.089619310061,0.046)
(0.44352280477,0.07)
(0.273297610244,0.058)
(0.29626110723,0.059)
(0.268533716269,0.067)
(0.08955670861,0.046)
(0.507175905478,0.07)
(0.313196838586,0.062)
(0.316819530336,0.06)
};

\addlegendentry{zh-en}
\addplot[draw=none,mark=pentagon,purple] coordinates {
(0.224487413284,0.01)
(0.1720901633,0.01)
(0.440665841775,0.008)
(0.24736565154,0.015)
(0.278632139752,0.011)
(0.247228769076,0.012)
(0.171916649768,0.01)
(0.470148267961,0.014)
(0.268314353623,0.008)
(0.296945787727,0.01)
(0.261885802325,0.015)
(0.172069038648,0.011)
(0.495103677298,0.018)
(0.293992413696,-0.002)
(0.312384898896,0.009)
(0.280121559093,0.02)
(0.172061551824,0.011)
(0.537280736582,0.018)
(0.332161613358,-0.009)
(0.325883699207,0.008)
};

\addplot[thick,blue] {0.0688 * x + 0.0232};
\addplot[thick,red] {-0.0171 * x + 0.0025};
\addplot[thick,black] {-0.0239 * x + 0.0029};
\addplot[thick,gray] {0.037 * x + 0.0105};
\addplot[thick,olive]{0.0225 * x + 0.0117};
\addplot[thick,orange] {0.0608 * x + 0.0413};
\addplot[thick,purple] {0.0101 * x + 0.0073};

\end{axis}
\end{tikzpicture}}
\resizebox{\linewidth}{!}{
\pgfplotsset{
  every axis plot/.append style={line width=0.9pt}
}
\begin{tikzpicture}

	\begin{axis}[
		height=6.6cm,
		width=\linewidth,
		grid=major,
		xmin=0,
		xmax=0.5,
		ylabel = Absolute gain in Kendall's $\tau$,
		xlabel = Diversity (DS$_{\textit{BOW}}$), 
		xtick distance = 0.1,
        axis x line*=bottom,
        axis y line*=left,
        legend style={draw=none,
            at={(1.4,1), font=\small,
            /tikz/every even column/.append style={column sep=0.9cm}},
    },
    legend columns=1
	]

\addlegendentry{de-en}
\addplot[draw=none,mark=triangle,blue] coordinates {
(0.185709145288,0.003)
(0.0746543504557,0.001)
(0.364103919534,0.005)
(0.192145472612,0.007)
(0.239330620017,0.005)
(0.216712020526,0.006)
(0.0741966857501,0.001)
(0.406292698522,0.018)
(0.230737241657,0.007)
(0.269963525226,0.01)
(0.239071851112,0.007)
(0.0742807581059,0.0)
(0.449424123242,0.025)
(0.264853692762,0.006)
(0.292012329416,0.005)
(0.265442506004,0.007)
(0.0743441329959,0.001)
(0.513844607653,0.023)
(0.305316505046,0.005)
(0.312841386936,0.008)
};

\addlegendentry{fi-en}
\addplot[draw=none,mark=square,red] coordinates {
(0.179380070839,0.02)
(0.067224397588,0.012)
(0.352168930985,0.014)
(0.186568123227,0.014)
(0.239742247385,0.014)
(0.214855945195,0.022)
(0.0667466038154,0.012)
(0.398914866529,0.023)
(0.223827903416,0.017)
(0.270567680591,0.019)
(0.237574420991,0.021)
(0.0667741075469,0.012)
(0.442896972763,0.028)
(0.261719190829,0.014)
(0.294236144162,0.021)
(0.263789441628,0.019)
(0.0668371802924,0.012)
(0.510841138874,0.026)
(0.304435572691,0.017)
(0.315291790801,0.028)
};

\addlegendentry{gu-en}
\addplot[draw=none,mark=o,black] coordinates {
(0.181524524533,0.016)
(0.0645037837645,0.009)
(0.320859016606,0.014)
(0.181932641431,0.013)
(0.249056935588,0.015)
(0.214016092374,0.014)
(0.0651720400453,0.009)
(0.372088694688,0.017)
(0.215881014343,0.012)
(0.270465321443,0.02)
(0.238584002409,0.013)
(0.0654352418899,0.01)
(0.417351869903,0.019)
(0.253755360147,0.016)
(0.292779656384,0.02)
(0.267272126295,0.018)
(0.0654990100256,0.01)
(0.485675344052,0.019)
(0.295277322229,0.011)
(0.31623493059,0.02)
};

\addlegendentry{kk-en}
\addplot[draw=none,mark=*,gray] coordinates {
(0.19394690794,0.001)
(0.0901129692546,0.005)
(0.35714718999,0.002)
(0.200729967661,-0.001)
(0.266323372536,0.011)
(0.226335427159,0.003)
(0.0905515814413,0.006)
(0.410110409491,0.006)
(0.241781899105,0.004)
(0.291589039057,0.008)
(0.250249134506,0.003)
(0.0902164072461,0.006)
(0.454733560258,0.01)
(0.274867126437,0.002)
(0.31337100985,0.007)
(0.275842935874,0.008)
(0.0899433525163,0.006)
(0.517551855,0.011)
(0.314518868276,0.005)
(0.333436608627,0.005)
};

\addlegendentry{lt-en}
\addplot[draw=none,mark=|,olive] coordinates {
(0.192409268255,0.025)
(0.114189104179,0.019)
(0.38090873517,0.034)
(0.211635964415,0.027)
(0.250666208805,0.027)
(0.223546688363,0.03)
(0.114874816275,0.018)
(0.418017365479,0.044)
(0.238271561085,0.033)
(0.276328809372,0.03)
(0.243013833544,0.035)
(0.115147556036,0.018)
(0.45267081486,0.052)
(0.26817711884,0.031)
(0.295758933091,0.036)
(0.263703756824,0.037)
(0.115382821325,0.019)
(0.507564808455,0.058)
(0.304980762987,0.039)
(0.313283537107,0.034)
};

\addlegendentry{ru-en}
\addplot[draw=none,mark=diamond,orange] coordinates {
(0.192356388546,0.014)
(0.0893812728579,0.008)
(0.357841406864,0.008)
(0.202511878767,0.008)
(0.251704295321,0.006)
(0.222860223282,0.015)
(0.089680914947,0.008)
(0.401354205522,0.013)
(0.237501374569,0.009)
(0.275047806255,0.005)
(0.243728566391,0.006)
(0.089619310061,0.008)
(0.44352280477,0.016)
(0.273297610244,0.012)
(0.29626110723,0.007)
(0.268533716269,0.008)
(0.08955670861,0.008)
(0.507175905478,0.013)
(0.313196838586,0.014)
(0.316819530336,0.007)
};

\addlegendentry{zh-en}
\addplot[draw=none,mark=pentagon,purple] coordinates {
(0.224487413284,-0.004)
(0.1720901633,0.0)
(0.440665841775,-0.007)
(0.24736565154,-0.001)
(0.278632139752,-0.007)
(0.247228769076,0.0)
(0.171916649768,-0.002)
(0.470148267961,-0.004)
(0.268314353623,-0.002)
(0.296945787727,-0.003)
(0.261885802325,0.002)
(0.172069038648,-0.002)
(0.495103677298,0.0)
(0.293992413696,-0.008)
(0.312384898896,-0.003)
(0.280121559093,0.003)
(0.172061551824,-0.001)
(0.537280736582,0.001)
(0.332161613358,-0.013)
(0.325883699207,-0.007)
};

\addplot[thick,blue] {0.0481 * x - 0.0046};
\addplot[thick,red] {0.0323 * x + 0.0102};
\addplot[thick,black] {0.0253 * x + 0.0086};
\addplot[thick,gray] {0.0103 * x + 0.0027};
\addplot[thick,olive]{0.0908 * x + 0.0082};
\addplot[thick,orange] {0.0119 * x + 0.0066};
\addplot[thick,purple] {-0.0041 * x - 0.0017};

\end{axis}
\end{tikzpicture}}
\caption{\label{fig:div-vs-corr}Lexical diversity versus absolute correlation gain at the system level (top) and segment level (bottom) for a variety of paraphrase systems (+2, +5, +10 and +20 references).}
\end{figure}
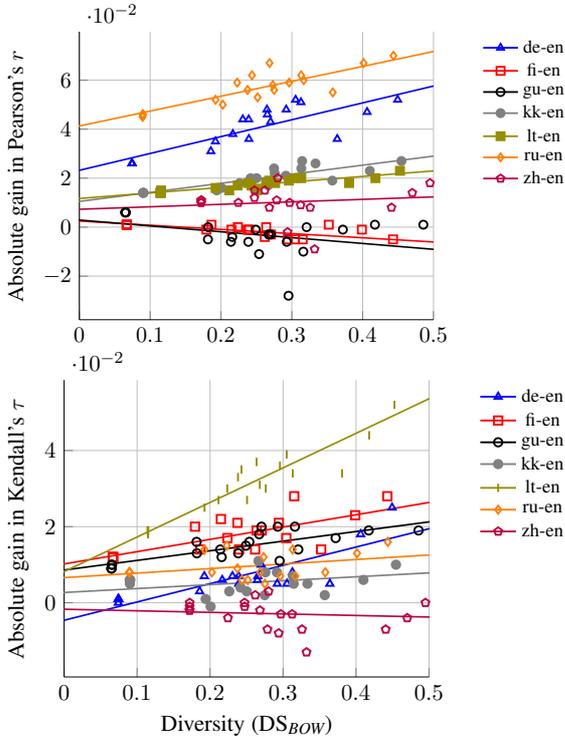

\rachel{Need to redo Figure 1 with absolute gains!}
Fig.~\ref{fig:div-vs-corr} (top) shows that for the diverse
approaches, the average 
gain is positively correlated with the method's diversity: increased
diversity does improve coverage of the valid translation space.  This
positive correlation holds for all directions for which adding
paraphrases helps (i.e.,~all except fi--en and gu--en).  For these
exceptions, none of the methods significantly improves over the
baseline, and \random gives as good if not marginally better results.
\rachel{Check this with redone results} The constraints approach
achieves the highest average 
gain, suggesting that it is more efficiently targeting the space of
valid translations, even though its paraphrases are significantly less
diverse (Tab.~\ref{tab:paraphrase_diversity-total-ave}).

Finally, and in spite of these improvements, we note that all systems
fall far short of the best WMT19 metrics, shown in the last row.
Automatic paraphrases do not seem to address the weakness of BLEU as
an automatic metric.

\paragraph{Segment-level} 

Similar results can be seen at the segment level, with most diverse
approaches showing improvements over the baseline (this time
\sentbleu) and a minority showing non-significant deteriorations
(i.e.,~no change). The diversity of the approaches is again positively
correlated with the gains seen (Fig.~\ref{fig:div-vs-corr}, bottom),
with the exception of zh--en, for no easily discernable reason.

The best result of the diverse approaches is again achieved by the
\sampled baseline.

The constraint-based approach achieves good scores, comparable to
\sampled, despite an anomalously poor score for one language pair (for
kk--en, with a degradation of 0.097.  This approach also had the
highest \bleu scores, however, suggesting that the targeted
paraphrasing approach here missed its mark.

\paragraph{De--en 500-sentence subset}

The general pattern shows the same as the averages over all languages
in Tab.~\ref{tab:correl-all}, with the more diverse methods
(especially \sampled) resulting in the greatest gains. The human
results also follow this pattern, resulting in the highest gains of
all at the system level.  Interestingly, the constrained system yields
higher average \bleu scores than \textsc{Human}
(Tab.~\ref{tab:paraphrase_diversity-total-ave}) yet a comparable
system correlation gain, indicating it targets more of the invalid
translation space.  For this particular subset, the constraints-based
approach helps slightly more at the segment level than the system
level, even surpassing the human paraphrases in terms of relative
gains, despite it having remarkably less
diversity.  

\section{Discussion}\label{sec:discussion}

\paragraph{Does diversity help?}
In situations where adding paraphrases helps (which is the case for a majority of language directions), the diversity of those paraphrases tends to positively correlate with gains in metric performance for both \bleu and \sentbleu.  
The adequacy of the paraphrases appears to be a less important factor, shown by the fact that the best automatic diverse method at both levels was the \sampled baseline, the most diverse but the least adequate.\footnote{We did not categorize our adequacy judgments, but \sampled's lower adequacy could be caused by (the relatively harmless) deletion of information 
(anecdotally supported in Tab.~\ref{tab:examples}).} 
The comparison against human paraphrases on the de--en subsample suggests room for improvement in automated techniques, at least at the system level, where all automatic metrics are beaten by \human paraphrases, which are both more diverse and more adequate.

However, diversity is not everything; although \human has nearly twice the lexical diversity of \sampled, it improves \bleu only somewhat and harms sentence \bleu.
On the other side, targeted constraints have relatively low diversity, but higher correlation gains.
Diversity itself does not necessarily result in coverage of the space occupied by good translation hypotheses.

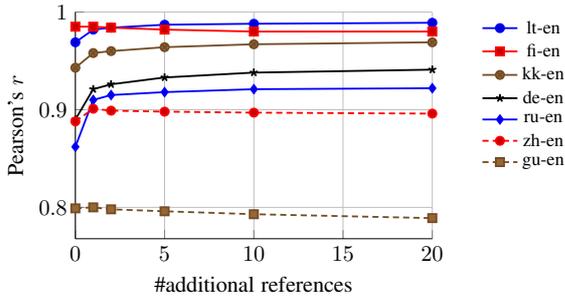
\begin{figure}[!ht]
\resizebox{\linewidth}{!}{
\pgfplotsset{
  every axis plot/.append style={line width=0.9pt}
}
\begin{tikzpicture}

	\begin{axis}[
		height=5.5cm,
		width=\linewidth,
		grid=major,
		xmin=0,
		xmax=20,
		ymax=1,
		ylabel = Pearson's $r$,
		xlabel = \#additional references, 
		xtick distance = 5,
		legend style={
            at={(1,-0.2)},
            legend columns=3,
            draw=none},
        axis x line*=bottom,
        axis y line*=left,
        legend style={draw=none,
            at={(1.4,1), font=\small,
            /tikz/every even column/.append style={column sep=0.9cm}},
    },
    legend columns=1
	]
	
	  \addlegendentry{lt-en}
\addplot coordinates {
 (0,0.969) (1,0.982) (2, 0.984) (5, 0.987) (10, 0.988) (20, 0.989)};
	
	 \addlegendentry{fi-en}
\addplot coordinates {
(0, 0.985) (1, 0.985) (2, 0.984) (5, 0.982) (10, 0.980) (20, 0.980)};

 \addlegendentry{kk-en}
\addplot coordinates {
 (0,0.943) (1,0.958) (2,0.960) (5,0.964) (10,0.967) (20,0.969)};
 
  \addlegendentry{de-en}
\addplot coordinates {
(0, 0.890) (1, 0.921) (2, 0.926) (5, 0.933) (10, 0.938) (20, 0.941)};
 
   \addlegendentry{ru-en}
\addplot coordinates {
 (0, 0.862) (1, 0.910) (2, 0.915) (5, 0.918) (10, 0.921) (20, 0.922)};

\addlegendentry{zh-en}
\addplot coordinates {
(0, 0.888) (1, 0.901) (2, 0.899) (5, 0.898) (10, 0.897) (20, 0.896)};
 
   \addlegendentry{gu-en}
\addplot coordinates {
(0, 0.799) (1, 0.800) (2, 0.798) (5, 0.796) (10, 0.793) (20, 0.789)};
 
\end{axis}
\end{tikzpicture}}
\vspace{-14pt}
\caption{\label{fig:treelstm-plain-num-sys}  \treelstmplain  system-level
correlations (+0-20).} 
\end{figure}

\paragraph{What effect do more references have?} Diversity increases the more paraphrases there are and it is positively correlated with gains for most language directions. However, improvements are slight, especially with respect to what we would hope to achieve (using human references results in much more diversity and also greater improvements).
The relationship between the number of extra references and system-level correlations shown in  
Fig.~\ref{fig:treelstm-plain-num-sys} suggests that increasing the number of references results in gains, but for most test sets, the initial paraphrase has the most impact and the subsequent ones
lead to lesser gains or even occasional deteriorations.
Similar results are seen at the segment level.

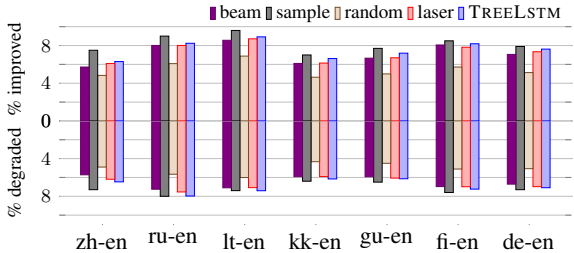
\begin{figure}[!ht]
\scalebox{0.8}{
\begin{tikzpicture}
  \begin{axis}[
    height=3.3cm,
    width=10cm,
    minor y tick num=1,
    yminorgrids,
    x dir=reverse,
    ymajorgrids,
    ytick distance=4,
    minor y grid style={line        width=.2pt,draw=gray!50},
    name=negative2,
    every y tick label/.append style={font=\small},
    ybar=-0pt,
    ylabel={\small\% degraded},
    ymin=0, ymax=11,
    y dir=reverse,
    bar width=-4pt,
    x axis line style = { opacity = 0 },
    tickwidth         = 0pt,
    x axis line style={line width=4pt},
    symbolic x coords = {de-en, fi-en, gu-en, kk-en, lt-en, ru-en, zh-en},
    legend image code/.code={
        \draw [#1] (0cm,-0.1cm) rectangle (0.08cm,0.1cm);
    },
    legend cell align={left},
    legend style={draw=none,
            at={(1,2.2), font=\small,
            /tikz/every even column/.append style={column sep=1.7cm}},
    },
    legend columns=5,
    reverse legend=true
  ]

  \addlegendentry{\treelstmplain}
  \addplot coordinates { 
    (de-en,7.09) (fi-en,7.24) (gu-en,6.14) (kk-en,6.16) (lt-en,7.41) (ru-en,7.97) (zh-en,6.47) 
  };
  
  
  \addlegendentry{laser}
  \addplot coordinates { 
    (de-en,6.98) (fi-en,6.99) (gu-en,6.07) (kk-en,5.91) (lt-en,7.08) (ru-en,7.54) (zh-en,6.19) 
  };
  
  \addlegendentry{random}
  \addplot coordinates { 
    (de-en,5.07) (fi-en,5.11) (gu-en,4.5) (kk-en,4.33) (lt-en,6.01) (ru-en,5.66) (zh-en,4.9) 
  };
  
  \addlegendentry{sample}
  \addplot coordinates { 
    (de-en,7.3) (fi-en,7.6) (gu-en,6.5) (kk-en,6.4) (lt-en,7.4) (ru-en,8.0) (zh-en,7.3)
  };
  
  \addlegendentry{beam}
  \addplot coordinates { 
    (de-en,6.71) (fi-en,6.97) (gu-en,5.93) (kk-en,5.92) (lt-en,7.1) (ru-en,7.25) (zh-en,5.72) 
  };
  
\end{axis}
\begin{axis}[
    height=3.3cm,
    width=10cm,
    x dir=reverse,
    at=(negative2.north west),
    minor y tick num=1,
    yminorgrids,
    ymajorgrids,
    ytick distance=4,
    minor y grid style={line width=.2pt,draw=gray!50},
    every y tick label/.append style={font=\small}, 
    ybar=-0pt,
    xtick=\empty,
    ylabel={\small\% improved},
    ymin=0, ymax=11,
    bar width=-4pt,
    x axis line style = { opacity = 0 },
    tickwidth         = 0pt,
    x axis line style={line width=4pt},
    symbolic x coords = {de-en, fi-en, gu-en, kk-en, lt-en, ru-en, zh-en},
    legend image code/.code={
        \draw [#1] (0cm,-0.1cm) rectangle (0.6cm,0.1cm);
    },
    legend cell align={left},
    legend style={draw=none,
            at={(2,1)}
    },
  ]
 
  \addplot coordinates { 
  (de-en,7.62)(fi-en,8.19)(gu-en,7.19)(kk-en,6.62)(lt-en,8.92)(ru-en,8.24)(zh-en,6.3)
  };
  
  
  \addplot coordinates { 
  (de-en,7.34)(fi-en,7.82)(gu-en,6.7)(kk-en,6.14)(lt-en,8.71)(ru-en,8.01)(zh-en,6.09)
  };
  
  \addplot coordinates { 
  (de-en,5.13)(fi-en,5.72)(gu-en,4.99)(kk-en,4.64)(lt-en,6.88)(ru-en,6.08)(zh-en,4.83)
  };
  
  \addplot coordinates { 
    (de-en,7.9) (fi-en,8.5) (gu-en,7.7) (kk-en,7.0) (lt-en,9.6) (ru-en,9.0) (zh-en,7.5)
  };
  
  \addplot coordinates { 
    (de-en,7.06)(fi-en,8.06)(gu-en,6.66)(kk-en,6.1)(lt-en,8.56)(ru-en,8.01)(zh-en,5.72)
  };
  
  \end{axis}
\end{tikzpicture}}
\caption{\label{fig:improvements-vs-degradations}\% improved and degraded (with respect to single-reference sentence-\bleu) for methods with +5 references. 
}
\end{figure}
\paragraph{Why are gains only slight?} 
With respect to the \sentbleu baseline, we calculate the percentage of comparisons for which the decision is improved (the baseline scored the worse translation higher than the better one and the new paraphrase-augmented metric reversed this)\footnote{`Better' and `worse' systems are determined by the official DA human assessments of MT quality.} and for which the decision is degraded (opposite reversal). 
The results (Fig.~\ref{fig:improvements-vs-degradations}) show that although all the systems improve a fair number of comparisons (up to 9.6\%), they degrade almost as many. So, while paraphrasing adds references that represent the space of valid translations, references are indeed being added that match with the space of invalid ones too.
Interestingly, the same pattern can be seen for human paraphrases, 6.46\% of comparisons being degraded vs.~8.30\% improved, suggesting that even when gold standard paraphrases are produced, the way in which the references are used by \sentbleu still rewards some invalid translations, though the balance is  shifted slightly in favour of valid translations. This suggests that at least at the segment level, \bleu is a balancing act between rewarding valid translations and avoiding rewarding invalid ones. Some of these effects may be smoothed out in system-level \bleu but there is still likely to be an effect. It is worth noting that for the two languages directions, fi--en and gu--en, for which diversity was negatively correlated with correlation gain (i.e., diversity could be harming performance), the most conservative approach (\random) leads to some of the best results.

\paragraph{What is the effect on individual $n$-grams?}

We study which new $n$-grams are being matched by the additional references for the two language directions with the largest system-level correlation gain (ru--en and de--en). 
For each sentence, we collect and count the $n$-grams that were not in the original reference but where in the five paraphrased references of \beam (missing \ngrams),\footnote{Using sacre\bleu's default \texttt{v13a} tokenization.}
accumulated across all test set sentences.
We also looked at the most frequent $n$-grams not found at all, even with the help of the paraphrases (i.e., the \emph{unrewarded $n$-grams} from Sec.~\ref{sec:contrained-paraphrasing}).
The results are in Tab~\ref{table:ngrams}.

\begin{table*}[!t]
\centering\small
\resizebox{\linewidth}{!}{
\begin{tabular}{rp{3in}p{3in}}
\toprule
    N & newly matched ngrams & missing ngrams \\
\midrule
1 
& a (494) of (480) , (442) to (370) in (364) The (315) the (273) is (204) for (196) has (196) on (193) was (179) have (171) that (166) be (155) at (145) been (140) with (138) and (134) 
& to (921) in (921) on (870) is (802) of (798) a (786) for (568) The (556) with (509) it (508) has (505) are (482) by (480) was (478) have (449) - (443) at (437) as (426) which (386) 
\\
\cmidrule{1-3}
4 
& U . S . (63) the U . S (39) , as well as (19) p . m . (15) for the first time (13) in accordance with the (12) the United States , (11) in the United States (10) a member of the (10) of the United States (9) The U . S (9) . m . on (9) , in order to (9) the United States and (8) , of course , (8) . S . Navy (8) . m . , (8) the Chinese Academy of (8) Chinese Academy of Engineering (8) the renaming of the (7)
& U . S . (136) , according to the (99) , " he said (77) the U . S (55) of the United States (48) of the Ministry of (39) the end of the (38) , " said the (37) same time , the (36) , such as the (36) as well as the (35) ( Xinhua ) -- (34) and so on . (33) , he said . (32) the head of the (32) , the head of (31) , as well as (30) on the basis of (30) , and so on (29) \\
\bottomrule
\end{tabular}}
\caption{Most frequently newly matched and missing $n$-grams for the de--en and ru--en test sets for \beam (+5).}
\label{table:ngrams}

\end{table*}

Unsurprisingly, most 1-grams are common grammatical words (e.g., \emph{a}, \emph{of}, \emph{to}, \emph{in}, \emph{the}) that may be present (or not) in any sentence; it is hard to draw any conclusions. 
For 4-grams, however, we see some interesting patterns.
Present in both lists are acronym variants such as \emph{U~.~S~.} for `United States' and \emph{p~.~m~.} for `afternoon' or the 24-hour clock; their presence on both sides indicates success in sometimes grabbing this variant as well as failure to do so consistently.
We also see phrasal variants such as \emph{,~according~to} and \emph{,~"~he~said}.
These last points corroborate a point made by \citet[\S7.2]{freitag-etal-2020-bleu} that references may omit these common variants.
It also suggests a more focused method for generating paraphrases: identify a high-precision set of common variants, and ensure their presence in the set of references, via constrained decoding or other means (in the spirit of Meteor's \citep{denkowski-lavie-2011-meteor} synonym-based matching). We note however, that our paraphrasing methods do seem to contain complementary information as they also tend to improve Meteor too (see results in ~App.~\ref{app:meteor-results}).

\section{Conclusion}
We studied the feasibility of using diverse automatic paraphrasing of English references to improve \bleu. Although increased diversity of paraphrases does lead to increased gains in correlation with human judgments at both the system and segment levels, the gains are small and inconsistent.
We can do a slightly better job by using cues from the system outputs themselves to produce paraphrases providing a helpful form of ``targeted'' diversity.
The comparison with manually produced paraphrases shows that there is room for improvement, both in terms of how much diversity is achieved and how much \bleu can be improved.
However, the lack of any improvement in some languages points to how hard it is to target this ``right kind'' of diversity \emph{a priori}; this, together with the relatively limited gains overall (especially in comparison with the best WMT19 metrics), suggests an intrinsic limit to \bleu's capacity to handle multiple references.

\section*{Acknowledgements}
We would like to thank MT Marathon 2019, during which this work was initiated and which also provide us with initial computing resources.
This work was supported by funding from the European Union's Horizon 2020 research and innovation programme under grant agreements No 825299 (GoURMET), 825303 and the UK Engineering and Physical Sciences Research Council (EPSRC) fellowship grant EP/S001271/1 (MTStretch).

\bibliography{anthology,custom}
\bibliographystyle{acl_natbib}

\clearpage
\appendix

\section{Number of distinct parse trees at different depths}\label{app:parse-depths}

Table~\ref{tab:numtrees} shows the number of distinct pruned tree at different depths. We choose a depth of 4 for the syntactic sentence encoding methods in our experiments.

\begin{table}[!ht]
  \centering\small
  \begin{tabular}{ rrrr }
    \toprule  
    depth & no leaves & type/token & with leaves \\ 
    \midrule
    1 & 16         & 0\%     & 16 \\ 
    2 & 207,794    & 1.0\%   & 207,794 \\
    3 & 2,158,114  & 11.2\%  & 2,629,907 \\
    4 & 6,089,874  & 31.6\%  & 10,631,249 \\
    5 & 8,865,720  & 46.1\%      & 14,102,645 \\
    \midrule
    $\infty$ & 13,054,272 & 68.1\% &  17,362,448\\
    \bottomrule
  \end{tabular}
  \caption{\label{tab:numtrees}Number of distinct pruned trees in different depths with and without leaves in the parsed data.}
\end{table}

\section{Paraphraser training details}\label{app:training-details}

All paraphrase models are Transformer base models \citep{vaswani_attention_2017}: 6 layers, 8 heads, word embedding dimension of 512, feedforward dimension of 2048. We set dropout to 0.1 and tie all embeddings to the output layer with a shared vocabulary size of 33,152. 
We use the same vocabulary (including the 256 cluster codes) for all models. 
We adopt Adam optimisation with a scheduled learning rate (initial $3\times10^{-4}$) and mini-batch size of 64. We train each model on 4 GTX Titan X GPUs with a gradient update delay of 2, and select the final model based on validation \bleu. 

\section{Sentence clustering training details}\label{app:cluster-training-details}

We set $k$ to 256 for $k$-means clustering. We train \treelstm sentence encoders using Adagrad with a learning rate of 0.025, weight decay of $10^{-4}$ and batch size of 400 for a maximum of 20 iterations. We set the model size to 256 and limit the maximum number of child nodes to 10.

\section{Full raw WMT19 results }\label{app:full-wmt19-results}

Table~\ref{tab:raw-wmt19-results} shows the raw correlations of each each paraphrase-augmented \bleu metric on WMT19 (system-level results top and segment-level results bottom). These correspond to the raw scores used to calculate the gains of each method with respect to the true baseline (\bleu or sentence\bleu) shown in the main results section in Table~\ref{tab:summary-correlations}.
We indicate the best system from WMT19 as a point of reference.

\section{Raw results for the de--en 500-sentence subset}

\begin{table}[!ht]

\centering\small
\resizebox{\linewidth}{!}{
\begin{tabular}{llll}
\toprule
&& \multicolumn{2}{c}{Correlation} \\
& Method & System & Segment \\
\midrule
Baseline & (sentence)\bleu & {0.895}   &  {0.026}  \\
\midrule
\multirow{3}{*}{\pbox{1.3cm}{Baselines (+5)}}
& \beam &  \textbf{0.934}   &  {0.048}  \\
& \random &  \textbf{0.926}   &  {0.043}  \\
& \sampled &  \textbf{0.939}   &  {0.069}  \\
\midrule
\multirow{2}{*}{\pbox{1.3cm}{Diversity (+1)}}
& \laser &  \textbf{0.929}   &  {0.048}  \\
& \treelstm &  \textbf{0.926}   &  {0.037}  \\
\midrule
\multirow{2}{*}{\pbox{1.3cm}{Diversity (+5)}}
& \laser &  \textbf{0.935}   &  {0.049}  \\
& \treelstm &  \textbf{0.939}   &  {0.034}  \\
\midrule
\midrule
\multirow{1}{*}{Constraints}
& 4-gram & \textbf{0.933} & 0.064 \\
\midrule
Human &&  \textbf{0.948}   &  {0.063}  \\
\bottomrule
\end{tabular}}
\caption{Correlations on the 500-sentence subset.}
\label{tab:subset-seg-level}
\end{table}

\begin{table*}[!ht]
\begin{subtable}{\linewidth}
\centering\small
\scalebox{0.97}{
\begin{tabular}{lllllllllll}
\toprule
 && de-en & fi-en & gu-en & kk-en & lt-en & ru-en & zh-en & Ave \\
 Approach & Method & (16) & (12) & (12) & (11) & (11) & (14) & (15)\\
\midrule
Baseline & \bleu &  {0.890} & {0.985} & {0.799} & {0.943} & {0.969} & {0.862} & {0.888} & 0.905 \\
\midrule
\multirow{3}{*}{\pbox{1.5cm}{Paraphrase baselines (+5)}}
 & \beam  & \textbf{0.928} & {0.984} & {0.793} & \textbf{0.961} & \textbf{0.986} & \textbf{0.921} & {0.900} & 0.925 \\
 & \random  & \textbf{0.916} & {0.986} & {0.805} & \textbf{0.957} & \textbf{0.983} & \textbf{0.908} & {0.898} & 0.922 \\
 & \sampled  & \textbf{0.937} & {0.984} & {0.798} & {0.966} & \textbf{0.989} & \textbf{0.929} & {0.902} & 0.929 \\
\midrule
\multirow{2}{*}{\pbox{1.5cm}{Diversity (+1)}}
 & \laser  & \textbf{0.919} & {0.987} & {0.799} & \textbf{0.957} & \textbf{0.981} & \textbf{0.909} & {0.904} & 0.922 \\
 & \treelstm  & \textbf{0.921} & {0.985} & {0.800} & {0.958} & \textbf{0.982} & \textbf{0.910} & {0.901} & 0.922 \\
\midrule
\multirow{2}{*}{\pbox{1.5cm}{Diversity (+5)}}
 & \laser  & \textbf{0.934} & {0.985} & {0.795} & \textbf{0.963} & \textbf{0.987} & \textbf{0.918} & {0.896} & 0.925 \\
 & \treelstm  & \textbf{0.933} & {0.982} & {0.796} & {0.964} & \textbf{0.987} & \textbf{0.918} & {0.898} & 0.925 \\
\midrule
Constraints & 4-grams  & \textbf{0.922} & 0.983 & 0.809 & \textbf{0.963} & \textbf{0.989} & \textbf{0.924} & \textbf{0.921} & 0.930\\
\midrule
& WMT-19 best  & \textbf{0.950}** & \textbf{0.995} & \textbf{0.993}*** & \textbf{0.998}*** & \textbf{0.989}* & \textbf{0.979}** & \textbf{0.988}*** & 0.985 \\
& & \tiny (\textsc{YiSi-1\_SRL}) & \tiny (\textsc{METEOR}) & \tiny (\textsc{YiSi-0}) & \tiny (\textsc{WMDO}) & \tiny (\textsc{ESIM}) & \tiny (\textsc{YiSi-1}) & \tiny (\textsc{ESIM}) \\
\bottomrule
\end{tabular}}

\caption{Pearson correlations at the system level.}
\vspace{0.3cm}
\end{subtable}
%
\begin{subtable}{\linewidth}
\centering\small
\scalebox{0.97}{
\centering\small
\begin{tabular}{lllllllllll}
\toprule
 && de-en & fi-en & gu-en & kk-en & lt-en & ru-en & zh-en & Ave \\
 Approach & Method & (32000) & (23952) & (12192) & (11000) & (11000) & (28000) & (30000)\\
\midrule
Baseline & sentence\bleu &  {0.055} & {0.228} & {0.175} & {0.368} & {0.251} & {0.114} & {0.317} & 0.215 \\
\midrule
\multirow{3}{*}{\pbox{1.5cm}{Paraphrase baselines (+5)}}
 & \beam  & {0.061} & {0.250} & {0.189} & {0.371} & {0.281} & {0.129} & {0.317} & 0.228 \\
 & \random  & {0.056} & {0.240} & {0.184} & {0.374} & {0.269} & {0.122} & {0.315} & 0.223 \\
 & \sampled  & {0.073} & {0.251} & {0.192} & {0.374} & {0.295} & {0.127} & {0.313} & 0.232 \\
\midrule
\multirow{2}{*}{\pbox{1.5cm}{Diversity (+1)}}
 & \laser  & {0.061} & {0.244} & {0.187} & {0.368} & {0.276} & {0.121} & {0.314} & 0.225 \\
 & \treelstm  & {0.061} & {0.242} & {0.185} & {0.383} & {0.278} & {0.123} & {0.315} & 0.227 \\
\midrule
\multirow{2}{*}{\pbox{1.5cm}{Diversity (+5)}}
 & \laser  & {0.062} & {0.245} & {0.187} & {0.372} & {0.284} & {0.123} & {0.315} & 0.227 \\
 & \treelstm  & {0.065} & {0.247} & {0.195} & {0.376} & {0.281} & {0.119} & {0.314} & 0.228 \\
\midrule
Constraints & 4-grams & 0.090 & 0.242 & 0.161 & 0.271 & 0.323 & 0.122 & 0.314 & 0.218 \\
\midrule
&  WMT-19 best  & \textbf{0.199}*** & \textbf{0.346}*** & \textbf{0.306}*** & \textbf{0.442}*** & \textbf{0.380}*** & \textbf{0.2\
22}*** & \textbf{0.431}*** & 0.333 \\
& &  \tiny(\textsc{YiSi-1$_{\text{SRL}}$}) &  \tiny\textsc{YiSi-1} &  \tiny(\textsc{YiSi-1}) & \tiny(\textsc{YiSi-1$_{\text{SRL}}$}) &  \tiny(\textsc{YiSi-1$_{\
\text{SRL}}$}) &  \tiny(\textsc{YiSi-1$_{\text{SRL}}$}) & \tiny(\textsc{YiSi-1$_\text{SRL}$}) \\
\bottomrule
\end{tabular}}

\caption{Kendall's $\tau$ at the segment level}
\end{subtable}
\caption{\label{tab:raw-wmt19-results}WMT19 correlations of paraphrased \bleu for each method against human assessments (\# judgments in brackets) .
Results that are significantly better than the sacre\bleu baseline are indicated as follows (at least $p\leq0.05$) are marked in bold.}
\end{table*}

\section{Results with the Meteor metric}\label{app:meteor-results}

Although we focus on ways of improving \bleu using paraphrases in this article, as \bleu is the dominant metric, it is also interesting to look at how adding paraphrases could help similar metrics. We apply the same method to improving the Meteor metric (version 1.5) \cite{denkowski:lavie:meteor-wmt:2014}, a metric which already integrates synonym support.

Summarised results (as gains with respect to the single-reference Meteor metric) are shown in Tab.~\ref{tab:parmeteor-summary-correlations} and 
raw results are shown in Tab.~\ref{tab:raw-meteor-wmt19-results} for both system-level and segment-level correlations. We observe that the true baselines (Meteor and sentenceMeteor) are improved in both cases, possibly more so than \bleu and in different ways, showing that the information added by the paraphrases is complementary to the synonym support offered by Meteor.


\begin{table*}[tp]
\centering\small
\begin{subtable}{0.75\linewidth}
\centering\small
\scalebox{0.95}{
\begin{tabular}{llrrrrrrrr}
\toprule
&&& \multicolumn{3}{c}{System} && \multicolumn{3}{c}{Segment} \\
Approach & Method & \hphantom{oo} & Ave. & Min & Max & \hphantom{oo} & Ave. & Min & Max \\
\midrule
\multirow{3}{*}{\pbox{1.8cm}{Baselines (+5)}}
& \beam &&  0.012 & {0.002} & {0.036}  &&  0.016 & {0.007} & {0.027} \\
& \random &&  0.009 & {0.002} & {0.028}  &&  0.010 & {0.004} & {0.022} \\
& \sampled &&  0.013 & {0.002} & {0.038}  &&  0.018 & {0.009} & {0.031} \\
\midrule
\multirow{3}{*}{\pbox{1.8cm}{Diversity (+1)}}
& \laser &&  0.009 & {0.002} & {0.025}  &&  0.011 & {0.005} & {0.017} \\
& \treelstm &&  0.009 & {0.001} & {0.025}  &&  0.011 & {0.004} & {0.019} \\
\midrule
\multirow{2}{*}{\pbox{1.8cm}{Diversity (+5)}}
 & \laser &&  0.014 & {0.003} & {0.034}  &&  0.015 & {0.007} & {0.021} \\
 & \treelstm &&  0.015 & {0.002} & {0.039}  &&  0.016 & {0.008} & {0.030} \\
\midrule
\multirow{2}{*}{\pbox{1.8cm}{Output-specific (+1)}}
& \laser &&  0.007 & {0.000} & {0.020}  &&  0.009 & {0.003} & {0.018} \\
& \treelstm &&  0.010 & {0.002} & {0.020}  &&  0.013 & {0.004} & {0.021} \\
\midrule
Constraints & 4-grams && 0.004 & -0.050 & 0.027 && -0.002 & 0.043 & -0.084\\
\bottomrule
\end{tabular}}
\end{subtable}
\caption{\label{tab:parmeteor-summary-correlations}Absolute gains in correlation for paraphrased Meteor for WMT19 with respect to the Meteor baseline. Significant gains (except for averages) are marked in bold ($p\leq0.05$).} 
\end{table*}%

\begin{table*}[!htp]
\begin{subtable}{\linewidth}
\centering\small
\scalebox{0.97}{
\begin{tabular}{lllllllllll}
\toprule
 && de-en & fi-en & gu-en & kk-en & lt-en & ru-en & zh-en & Ave \\
 Approach & Method & (16) & (12) & (12) & (11) & (11) & (14) & (15\\
\midrule
Baseline & \textsc{Meteor} &  {0.909} & {0.993} & {0.883} & {0.969} & {0.972} & {0.825} & {0.941} & 0.927 \\
\midrule
\multirow{3}{*}{\pbox{1.9cm}{Paraphrase baselines (+5)}}
 & \beam  & \textbf{0.927} & {0.994} & {0.887} & {0.976} & \textbf{0.983} & {0.862} & \textbf{0.949} & 0.940 \\
 & \random  & \textbf{0.920} & {0.994} & {0.889} & {0.974} & \textbf{0.981} & {0.853} & {0.945} & 0.937 \\
 & \sampled  & \textbf{0.925} & {0.995} & {0.891} & {0.978} & \textbf{0.982} & {0.864} & {0.945} & 0.940 \\
\midrule
\multirow{2}{*}{\pbox{1.9cm}{Diversity (+1)}}
 & \laser  & \textbf{0.924} & {0.995} & {0.886} & \textbf{0.975} & \textbf{0.979} & {0.851} & \textbf{0.948} & 0.937 \\
 & \treelstm  & \textbf{0.923} & {0.994} & {0.889} & {0.974} & \textbf{0.979} & {0.850} & \textbf{0.947} & 0.937 \\
\midrule
\multirow{2}{*}{\pbox{1.9cm}{Diversity (+5)}}
 & \laser  & \textbf{0.932} & {0.995} & {0.890} & \textbf{0.978} & \textbf{0.983} & {0.860} & \textbf{0.950} & 0.941 \\
 & \treelstm  & \textbf{0.930} & {0.995} & {0.894} & {0.977} & \textbf{0.983} & {0.864} & \textbf{0.950} & 0.942 \\
\midrule
Constraints & 4-grams & 0.922 & 0.990 & 0.910 & 0.983 & 0.988 & 0.775 & 0.949 & 0.931 \\
\midrule
& WMT-19 best  & \textbf{0.950} & \textbf{0.995} & \textbf{0.993} & \textbf{0.998} & \textbf{0.989} & \textbf{0.979} & \textbf{0.988} & 0.985 \\
& & \tiny (\textsc{YiSi-1\_SRL}) & \tiny (\textsc{METEOR}) & \tiny (\textsc{YiSi-0}) & \tiny (\textsc{WMDO}) & \tiny (\textsc{ESIM}) & \tiny (\textsc{YiSi-1}) & \tiny (\textsc{ESIM}) \\
\bottomrule
\end{tabular}}

\caption{Pearson correlations at the system level.}
\vspace{0.3cm}
\end{subtable}
%

\begin{subtable}{\linewidth}
\centering\small
\scalebox{0.97}{
\begin{tabular}{lllllllllll}
\toprule
 && de-en & fi-en & gu-en & kk-en & lt-en & ru-en & zh-en & Ave \\
 Approach & Method & (32000) & (23952) & (12192) & (11000) & (11000) & (28000) & (30000\\
\midrule
Baseline & sentence\textsc{Meteor} &  {0.061} & {0.243} & {0.197} & {0.356} & {0.275} & {0.145} & {0.351} & 0.233 \\
\midrule
\multirow{3}{*}{\pbox{1.9cm}{Paraphrase baselines (+5)}}
 & \beam  & {0.081} & {0.257} & {0.219} & {0.383} & {0.285} & {0.152} & {0.360} & 0.248 \\
 & \random  & {0.072} & {0.254} & {0.219} & {0.364} & {0.281} & {0.156} & {0.356} & 0.243 \\
 & \sampled  & {0.080} & {0.262} & {0.228} & {0.375} & {0.292} & {0.160} & {0.360} & 0.251 \\
\midrule
\multirow{2}{*}{\pbox{1.9cm}{Diversity (+1)}}
 & \laser  & {0.079} & {0.258} & {0.209} & {0.370} & {0.283} & {0.150} & {0.359} & 0.244 \\
 & \treelstm  & {0.074} & {0.255} & {0.210} & {0.374} & {0.284} & {0.149} & {0.357} & 0.243 \\
\midrule
\multirow{2}{*}{\pbox{1.9cm}{Diversity (+5)}}
 & \laser  & {0.078} & {0.257} & {0.214} & {0.377} & {0.293} & {0.158} & {0.358} & 0.248 \\
 & \treelstm  & {0.074} & {0.259} & {0.228} & {0.378} & {0.287} & {0.153} & {0.361} & 0.249 \\
\midrule
Constraints & 4-grams & 0.098 & 0.237 & 0.193 & 0.272 & 0.318 & 0.145 & 0.351 & 0.230 \\
\midrule
&  WMT-19 best  & \textbf{0.20} & \textbf{0.35} & \textbf{0.31} & \textbf{0.44} & \textbf{0.38} & \textbf{0.22} & \textbf{0.43} & 0.333 \\
& & \tiny \tiny (\textsc{YiSi-1$_{\text{SRL}}$}) & \tiny (\textsc{YiSi-1}) & \tiny (\textsc{YiSi-1}) & \tiny (\textsc{YiSi-1$_{\text{SRL}}$}) & \tiny (\textsc{YiSi-1$_{\text{SRL}}$}) & \tiny (\textsc{YiSi-1$_{\text{SRL}}$}) & \tiny (\textsc{YiSi-1$_\text{SRL}$}) \\
\bottomrule
\end{tabular}}
\caption{Kendall's $\tau$ at the segment level}
\end{subtable}
\caption{\label{tab:raw-meteor-wmt19-results}WMT19 correlations of paraphrased \textsc{Meteor} for each method against human assessments (\# judgments in brackets) .
Results that are significantly better than the \textsc{Meteor} baseline are indicated as follows (at least $p\leq0.05$) are marked in bold.}
\end{table*}


\section{Further examples of automatically paraphrased references}\label{app:more-examples}
We provide additional examples of paraphrased references. As can be seen from Table \ref{tab:extra-paraphrases}, \treelstm gives us more diverse sentences compared to \laser.
\begin{table*}[ht!]
\centering\small
\begin{adjustbox}{totalheight=\textheight-2\baselineskip}
{
\begin{tabular}{lp{.9\textwidth}}
\toprule
Reference &  \textit{Anyway the application phase for each of the 20 properties, which the municipality is offering at better conditions and prices - similar to the earlier local model, has now ended.}  \\
\midrule
\multirow{3}{*}{\laser} 
  &	 However, the application phase for each of the 20 properties offered by the municipality at better conditions and prices, similar to the previous local model, is now over. \\
  &	 However, the application phase for each of the 20 properties offered by the municipality at better conditions and prices, similar to the previous local model, is now over. \\
  &	 However, the application phase for each of the 20 properties offered by the municipality at better conditions and prices, similar to the previous local model, is now over. \\
\midrule
\multirow{3}{*}{\treelstmplain}
 & However, the application phase for each of the 20 properties offered by the municipality at better conditions and prices - similar to the previous local model - is now closed. \\
  &	 In any case, the application phase for each of the 20 properties offered by the municipality at better conditions and prices - similar to the previous local model - has now come to an end. \\
  &	 However, for each of the 20 properties that the municipality offers at better conditions and prices, similar to the previous local model, the application phase has now come to an end. \\ 
\midrule
\midrule
Reference & \textit{And there is a need for discussion.} \\
\midrule
\multirow{3}{*}{\laser}
  &	 And we need to discuss it.\\
  &	 And we need to have a discussion. \\
  &	 And that's what we need to discuss. \\
 \midrule
 \multirow{3}{*}{\treelstmplain}
  &	 And there's a need to discuss it. \\
  &	 And we need to discuss it. \\
  &	 It is necessary to discuss it. \\ 
\midrule
\midrule
Reference & \textit{Why are there increasingly conflicts, despite successful integration?} \\
\midrule
\multirow{3}{*}{\laser}
&	 Why is there ever more conflict, despite successful integration? \\
  &	 Why is there ever more conflict, despite successful integration? \\
  &	 Why are there more and more conflicts, despite the success of integration? \\
 \midrule
 \multirow{3}{*}{\treelstmplain}
  &	 Why is there more and more conflict, despite successful integration? \\
  &	 Why are conflicts growing despite successful integration? \\ 
  &	 Why are there ever more conflicts, despite the success of integration? \\
\midrule
\midrule
Reference & \textit{Professor Doctor Aladin El-Mafaalani is standing before them.} \\
\midrule
\multirow{3}{*}{\laser}
  &	 In front of them is Professor Dr. Aladin El-Mafaalani. \\
  &	 Professor Dr. Aladdin El-Mafaalani stands in front of them. \\
  &	 Professor Dr. Aladdin El-Mafaalani stands in front of them. \\
 \midrule
 \multirow{3}{*}{\treelstmplain}
  &	 Professor Aladdin El-Mafaalani stands before them. \\
  &	 In front of them is Professor Aladdin El-Mafaalani. \\
  &	 There's Professor Aladdin El-Mafaalani in front of them. \\
\midrule
\midrule
Reference & \textit{However, the operation took some time.} \\
\midrule
\multirow{3}{*}{\laser}
 &	 But the surgery took a while. \\
  &	 But the operation took a while. \\
  &	 But the surgery took a while. \\
\midrule
\multirow{3}{*}{\treelstmplain} 
  &	 However, the surgery lasted a while. \\
  &	 But the surgery lasted a while. \\
  &	 However, it took some time to operate. \\
\midrule
\midrule
Reference & \textit{According to the police, it appears the fire was monitored for a long time, but the embers were inadvertently left behind.} \\
\midrule
\multirow{3}{*}{\laser}
  &	 According to the police, the fire seems to have been under long surveillance, but the coals were inadvertently left behind. \\
  &	 According to the police, the fire seems to have been under long surveillance, but the coals have been inadvertently left behind. \\
  &	 According to the police, the fire seems to have been under long surveillance, but the coals were inadvertently left behind. \\
\midrule
\multirow{3}{*}{\treelstmplain} 
&	 Police say the fire seems to have been under observation for a long time, but the coals are accidentally left behind. \\
  &	 The police seem to have been watching the fire for a long time, but the coals were accidentally left behind. \\
  &	 According to the police, the fire seems to have been under observation for a long time, but the coals were accidentally left behind. \\
\midrule
\midrule
Reference & \textit{What is the situation on the island now?} \\
\midrule
\multirow{3}{*}{\laser}
 &	 How's the island now? \\
  &	 What's happening on this island now? \\
  &	 What's the status on the island these days? \\
\midrule
\multirow{3}{*}{\treelstmplain} 
  &	 What's it like on the island? \\
  &	 What's going on on the island? \\
  &	 So what's the status on the island? \\

\bottomrule
\end{tabular}
}
\end{adjustbox}
\caption{Top three paraphrases for seven sentences.}
\label{tab:extra-paraphrases}
\end{table*}

\end{document}